\documentclass[letterpaper]{article} 
\usepackage{aaai}
\usepackage{times}
\usepackage{helvet}
\usepackage{courier}
\usepackage{xspace}
\usepackage{algorithm}
\usepackage{subcaption}
\usepackage[noend]{algorithmic}
\usepackage{graphicx} 
\usepackage{url}
\usepackage{comment}
\usepackage{multirow,multicol, makecell, booktabs}
\usepackage{amsmath}
\usepackage{amsthm}
\usepackage{amsfonts}
\usepackage{color}
\usepackage{lipsum}
\usepackage{listings}
\usepackage{todonotes}
\usepackage{pdflscape}
\newcommand{\shira}[1]{({\color{magenta}Shira: #1})}
\newcommand{\alberto}[1]{\textcolor{blue}{#1}}

\newcommand{\jpmc}{J.P. Morgan\xspace}
\newcommand{\expres}{\textsc{expres}\xspace}
\newcommand{\pres}{\textsc{pres}\xspace}

\newtheorem{definition}{Definition}
\newtheorem{theorem}{Theorem}

\begin{document}
\title{Explaining Preference-driven Schedules: the EXPRES Framework\\\normalsize{Extended Version}}

\author{Alberto Pozanco$^1$,  Francesca Mosca$^1$,  Parisa Zehtabi$^1$,  Daniele Magazzeni$^1$,  Sarit Kraus$^2$\\
$^1$J.P. Morgan AI Research\\
$^2$Department of Computer Science, Bar-Ilan University\\
\{alberto.pozancolancho, francesca.mosca, parisa.zehtabi, daniele.magazzeni\}@jpmorgan.com, sarit@cs.biu.ac.il
}

\maketitle 

\begin{abstract}
Scheduling is the task of assigning a set of scarce resources distributed over time to a set of agents, who typically have preferences about the assignments they would like to get.
Due to the constrained nature of these problems, satisfying all agents' preferences is often infeasible, which might lead to some agents not being happy with the resulting schedule.
Providing explanations has been shown to increase satisfaction and trust in solutions produced by AI tools. However, it is particularly challenging to explain solutions that are influenced by and impact on multiple agents.
%%
% However, explaining schedules poses some particular challenges, such as problem interpretability (i.e., generating explanations from a huge and dense amount of information) and privacy preservation (i.e., generating explanations respecting the privacy of other agents involved). 
In this paper we introduce the \expres framework, which can explain why a given preference was unsatisfied in a given optimal schedule. The \expres framework consists of: (i) an \textit{explanation generator} that, based on a Mixed-Integer Linear Programming model, finds the best set of reasons that can explain an unsatisfied preference; and (ii) an \textit{explanation parser}, which translates the generated explanations into human interpretable ones. %, while preserving agents' privacy.
Through simulations, we show that the explanation generator can efficiently scale to large instances.
Finally, through a set of user studies within \jpmc, we show that employees preferred the explanations generated by \expres over human-generated ones when considering workforce scheduling scenarios.
\end{abstract}

%%%%%%%%%%%%%%%%%%%%%%%%%%%%%%%%%%%%%%%%%%%%%%%%%%%%%%%%%%%%%%%%%%%%%%%%

\section{Introduction}
Scheduling is the task of assigning a set of scarce resources distributed over time to a set of agents. 
This is the case of many well-known problems, such as assigning jobs to machines~\cite{DBLP:journals/ai/WatsonBHW03}, nurses to work shifts~\cite{legrain2015nurse}, or teachers to courses~\cite{gunawan2007solving}, among others.
Another application that has become very relevant to organisations (including \jpmc) due to COVID-19 restrictions is that of scheduling or assigning employees to a limited number of desks (less than in normal situations in order to guarantee social distancing) over a fixed time period. In this context, employees may have specific preferences regarding their schedules, such as the dates or the number of days a week that they want to be at the workplace, or the peers they would like to meet more often, etc. 
However, the limited availability of desks may preclude the fulfillment of all of the employees' preferences, and this may hinder their satisfaction with the schedule. Presenting a contrastive explanation of the reasons why the employees could not be scheduled in any other way may promote the acceptance of the schedule \cite{bradley2009dealing}.
% In fact, providing explanations has been shown to be a valid way of increasing users' satisfaction and trust in solutions produced by AI tools~\cite{bradley2009dealing}.
Furthermore, beside the intrinsic difficulty of explaining a schedule for an individual, more challenges arise when considering the automated explanation of decisions regarding multiple agents. Among the features that Kraus \textit{et al.} (\citeyear{kraus2020ai}) argue for, an explainable system should (i) allow its users to understand its decision-making, (ii) be able to generate different types of explanations, so to provide tailored outputs for its users, (iii) while preserving the agents' privacy.
% Kraus \textit{et al.} (\citeyear{kraus2020ai}) identified three main challenges or research directions when explaining solutions to multiple agents: (1) user modeling for increasing satisfaction; (2) interactive explanations; and (3) development of efficient algorithms able to generate sets of explanations from which users can select.
% %%
% In this work we have addressed some of these challenges by proposing a framework that can quickly generate multiple different explanations given an unsatisfied preference selected by the user, \francesca{increasing their satisfaction with the resulting schedules.}

\begin{figure}
    \centering
    \includegraphics[width=.9\linewidth]{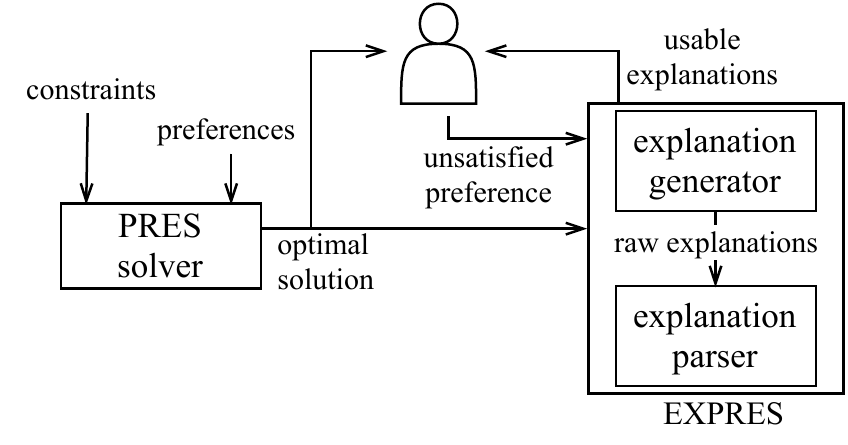}
    \caption{Overview of the \expres framework.}
    \label{fig:expres-overview}
\end{figure}

In this paper we present the \expres framework (see Fig.~\ref{fig:expres-overview}), a novel approach to providing EXplanations for PREference-driven Scheduling problems (\pres). 
Our formalisation is general, allowing it to be applied in different scenarios; however, for the sake of clarity, we discuss and empirically evaluate a complete example of \expres in the context of workforce scheduling at \jpmc.

We first define \pres problems as scheduling tasks where an optimal solution is identified not only by a set of constraints, but also by a totally ordered set of preferences.
We then formalise the problem of explaining \pres solutions (\expres) as an optimisation task where, given a schedule and an agent's unsatisfied preference, we find the best set of reasons that can justify it. 
On one hand, EXPRES explanations simplify the task, frequently manually performed with notable effort and time-consumption, of providing justification for a specific unsatisfied preference in a given schedule. On the other hand, these explanations aim to support the individual's understanding of the other factors, beyond their own preferences, that influenced the schedule. 
% not necessarily to increase the individual's satisfaction towards the schedule (that’s left as future work), as this may depend strongly on the problem constraints, or to improve the collaboration with the scheduler (i.e., they are not actionable), but
We propose to model \expres as a Mixed-integer Linear Programming (MILP) problem.
After that, we show how to group and translate the computer generated explanations to natural language in order for them to be easily interpreted by humans.
After discussing how to convey the explanations to end users, we show through software simulations that \expres is able to (i) scale to large instances, in terms of number of employees and preferences; and (ii) provide different explanations, in terms of reasons to justify the unsatisfied preference.
Later, we also present the results of a user study within \jpmc that shows how \expres explanations are better appreciated than human-generated ones. %\francesca{and increase employees' satisfaction and trust in the solutions produced by a workforce scheduling tool.}
Finally, we draw our main conclusions and outline future work.

\section{\pres: Preference-driven Scheduling}
In this section we formalise a scheduling problem where a finite set of \emph{resources} $R$ distributed across different \emph{time slots} $T$ needs to be assigned to a set of \emph{agents} $Ag$.
Each agent might have a set of constraints as well as preferences of different types.
An external actor, namely the Principal, specifies a set of global constraints and a total order over the agents' preference types\footnote{This problem setting was considered in a real-world scenario, namely the return to the office at \jpmc, as we discuss later.}.
%%
% This is the problem setting that was considered as a use case in the return to the office at \jpmc, as we will discuss later.
%%
We refer to this set of scheduling problems as \pres, and formally define them as follows:
% \begin{definition}
% A \textbf{\pres problem} is a tuple $\mbox{\pres}=\langle R, Ag, T, {\cal C}, P_{ag_1}, ..., P_{ag_{|Ag|}}, {\cal P}_\preceq \rangle$ where:
% \begin{itemize}
%     \item $R$ is a set of resource types
%     \item $Ag$ is a set of agents
%     \item $T$ is a set of time slots
%     \item $\cal C$ is a set of constraints
%     \item $ag_{P}$ is the set of preferences of agent $ag \in Ag$
%     % \item ${\cal P} = \bigcup_{ag \in Ag} P_{ag}$ is the set of all agents' preferences
%     \item ${\cal P_\preceq} = \{P_x, P_y,...  \}$ is a total order set of all the agents' preferences, according to their type $x, y .... \in X$ .
    
%     % $p_{ag,x} \in P_x$
% \end{itemize}
% \end{definition}

\begin{definition}
A \textbf{\pres problem} is a tuple $\mbox{\pres}=\langle R, Ag, T, {\cal C}, {\cal P}, {\cal O} \rangle$ where:
\begin{itemize}
    \item $R$ is a set of resource types
    \item $Ag$ is a set of agents
    \item $T$ is a set of time slots
    % \item $C_{ag}$ and $P_{ag}$ are respectively the constraints and the preferences of agent $ag \in Ag$
    \item ${\cal C} = \bigcup_{ag \in Ag} C_{ag} \cup C_{Princ}$ is the set of all the constraints, where $C_{ag}$ are the constraints of the agent $ag$, and $C_{Princ}$ are the constraints imposed by the Principal 
    \item ${\cal P} = \bigcup_{ag \in Ag} P_{ag}$ is the set of all agents' preferences
    % \item ${\cal P_\preceq} = \{P_x, P_y,...  \}$ is a total order set of all the agents' preferences, according to their type $x, y .... \in X$ .
    \item ${\cal O}=\{\tau_1 \prec \ldots \prec \tau_{|\cal{O}|} \}$ is a totally ordered set of preference types $\tau_i$ defined by the Principal; $\tau_i \prec \tau_j$ means that $\tau_i$ precedes (is more important than) $\tau_j$ in the order. 
    % \item $\preceq$ is a total order defined over the preference-types in $\mathcal{P}$: $P(x_1) \preceq ... \preceq P(x_{|X|})$
\end{itemize}
\end{definition}
Given a preference $p \in {\cal P}$, we refer to the agent having the preference, to its type and to the order of its type as $ag(p)$, $\tau(p)$ and $\mathcal{O}(p)$ respectively. 
The solution to a \pres problem is a \textbf{schedule} $S$, which consists of a set of \textbf{assignments} $as = \langle ag, r, t \rangle \in \{Ag \times R \times T\}$ that assign agents to resources and time slots. 
%%
%Given an assignment $as$, we refer to the assigned agent, the assigned resource and the assigned time-slot respectively with $as(ag)$, $as(r)$ and $as(t)$.
Again, $ag(as)$ and $r(as)$ refer to the agent and the resource of the assignment respectively.
We refer as ${\cal S} \subseteq \{Ag \times R \times T\}$ to the set of all \textbf{feasible schedules} subject to the constraints in $\cal C$.
The optimal solution to a \pres problem is defined according to the totally ordered set of preference types ${\cal O}$.
%%
% We denote the order in $\cal P$ of a preference-type with a superscript, e.g., $P_x^1 \subseteq \cal{P}$ means that $P_x$ is the most important preference-type in $\cal P$.
%%
% We denote the order in $\cal O$ of a preference with a superscript, e.g., $\tau(p)=1$ 
% means that $p$ is of the most important preference-type in $\cal O$.
%In order to formally express this, we introduce the following function:
\begin{definition}
Given a set of solutions $\tilde{\cal S}$ and a preference type $\tau$, a \textbf{preference filter function} $ f_{\tau}: \tilde{\cal S} \longrightarrow \tilde{\cal S}_{\tau} $ returns the set of solutions that maximises the number of preferences $p \in {\cal P}$ of type $\tau$ that are satisfied.
\end{definition}
We formally define an optimal schedule given a totally ordered set of preference types and the set of feasible schedules as follows:
\begin{definition}
A \textbf{set of optimal schedules} $\cal S^*$ is the output of a composition of preference filter functions, where the order of composition is given by the total order $\cal O$:
$$ {\cal S}^* = f_{\tau_{|O|}} \circ \cdots \circ f_{\tau_1} (\cal S) $$
where $\tau_1$ is the first (most important) preference type.
\end{definition} 
In the rest of the paper we assume that the computed solutions are optimal and that the agents' preferences do not contradict their own constraints or the constraints defined by the Principal (e.g., an agent cannot ask for a number of resources higher than their own maximum number or higher than the number of available resources).
% \skcomment{Not self-contradictory is not well defined. You do not refer to specific agent in the preferences. You just refer to preferences types. What are the preferences of a given agent?}

%%%%%%%%%%%%%%%%%%%%%%%%%%%%%%%%%%%%%

\section{\expres: Explaining \pres Solutions}
Given the constrained nature of the \pres problems, it may happen that in a schedule $S$ some agents' preferences are not accommodated.
% After solving a \pres problem and due to its constrained nature, some agents' preferences might not be accommodated in a schedule $S$.
%%
We refer to these as \textbf{unsatisfied preferences} $\mbox{UNSAT} \subseteq {\cal P}$, using $\mbox{SAT} \subseteq {\cal P}$ to denote the set of \textbf{satisfied preferences}. Note that $\mbox{UNSAT} \cap \mbox{SAT} = \emptyset$.
%%
%\skcomment{$SAT$ is overloaded.}
In such scenarios we aim to provide agents with informative explanations that clarify why some of their preferences were unsatisfied.
Particularly, we explain unsatisfied preferences by reporting the \emph{reasons} why the preferred assignments had to be assigned in a different way.
We refer to this task as \textit{explaining} \pres solutions (\expres); before formalising it, we define some concepts.

In order to provide an explanation for why preference $u \in \mbox{UNSAT}$ is unsatisfied, we first need to identify the set of assignments that were responsible for or involved in $u$ being unsatisfied.
%%
% Those will be the assignments we will need to justify in order to explain $u$.
\begin{definition}
$\textsc{Involved}(S,u)$ is a domain-specific function that computes the \textbf{set of assignments} $S^\prime \subseteq S$ \textbf{involved in} $u \in \mbox{UNSAT}$ being \textbf{unsatisfied}.
\end{definition}
For example, assuming an unsatisfied preference $u$, an assignment $as$ is involved in $u$ if $t(as) = t(u)$, i.e., if there is another agent assigned to the time slot to which agent $ag(u)$ preferred to be assigned.
The \textsc{Involved} function relates assignments to unsatisfied preferences, returning the set of assignments to be justified or explained.

We also need a way of relating satisfied preferences to assignments, i.e., if a given assignment is affected or not by a preference.
\begin{definition}
$\textsc{Affected}(p,as)$ is a domain-specific function that returns a binary output indicating whether \textbf{assignment} $as \in S$ \textbf{is affected by satisfied preference} $p \in \mbox{SAT}$.
\end{definition}
For example, assuming a satisfied preference $p$, an assignment $as$ is affected by $p$ if $ag(p)=ag(as)$, i.e., the agent $ag$ had a preference regarding that assignment.

Explanations in our setting are formed by reasons that justify why the preferred assignments had to be assigned in a different way.
We formally define a reason as follows:
\begin{definition}
Given a \pres problem and a schedule $S$ that solves it, a \textbf{reason} is a tuple ${\cal R}=\langle p, as, u \rangle$, where $p \in \mbox{SAT} \subseteq {\cal P}$ is a satisfied preference, $as \in S$ is an assignment, and $u  \in {\mbox{UNSAT}} \subseteq {\cal P}$ is an unsatisfied preference.
\end{definition}
Reasons can be read as ``preference $u$ about assignment $as$ was unsatisfied due to preference $p$ being satisfied".
\begin{definition}
\label{def:well_defined_reason}
A reason ${\cal R}=\langle p, as, u \rangle$ is \textbf{well defined} iff (1) $\textsc{Affected}(p,as) = 1$; (2) $as \in \textsc{Involved}(S,u)$; and (3) $\textsc{Rank}(p, \leq, u)$.
\end{definition}
Here we assume a \textsc{Rank} function that returns $1$ (True) if ${\cal O}(p) \leq {\cal O}(u)$.
%%
%We will also refer to \textsc{ranking}($p^y$) as a function that returns $y$, i.e., the position of the given preference in $\cal P$.
%%
A well defined reason employs a more important preference over an assignment to explain why a lower (or equally) important preference over that same assignment was unsatisfied.
If both preferences have the same rank (${\cal O}(p) = {\cal O}(u)$), the reason suggests that there exists another optimal schedule in which $u$ was satisfied, because they are equally important.
With these definitions at hand we are ready to formalise an \expres problem as follows:
\begin{definition}
An \textbf{\expres problem} is a tuple $\expres = \langle \pres, S, u \rangle$ where:
\begin{itemize}
    \item $\mbox{\pres}=\langle R, Ag, T, {\cal C}, {\cal P}, {\cal O}\rangle$ is a \pres problem
    
    \item $S$ is a set of assignments that optimally solve \pres
    
    \item $u \in {\mbox{UNSAT}}$ is an unsatisfied preference
\end{itemize}
The \textbf{solution to an \expres problem} is an explanation ${\cal E}_{S,u}$, which is a set of reasons that describe why a preference $u$ was unsatisfied in a schedule $S$ that optimally solves a \pres problem.
\end{definition}
Explanations have different properties depending on the reasons they contain.
We say that an explanation is complete if it provides a reason for each of the assignments involved in an unsatisfied preference.
\begin{definition}
\label{def:explanation_completeness}
An \textbf{explanation} $\mathcal{E}_{S,u}$ that solves an \expres problem is \textbf{complete} iff
        $\forall as \in \textsc{Involved}(S,u),\, \exists {\cal R} \in \mathcal{E}_{S,u} \mid as= as({\cal R}).$
\end{definition}
We also say that an explanation is sound if all of the reasons in the explanation are well-defined.
\begin{definition}
\label{def:explanation_soundness}
An \textbf{explanation} $\mathcal{E}_{S,u}$ that solves an \expres problem is \textbf{sound} iff ${\cal R}\, \mbox{is well-defined}\, \forall {\cal R} \in {\mathcal{E}_{S,u}}$.
\end{definition}
Finally, we say that an explanation is optimal if it minimizes the order of the satisfied preferences used by its reasons.
\begin{definition}
\label{def:explanation_optimality}
An \textbf{explanation} $\mathcal{E}_{S,u}$ that solves an \expres problem is \textbf{optimal} iff $\min_{{\cal O}(p)} \sum_{{\cal R} \in {\mathcal{E}_{S,u}}} {\cal O}(p({\cal R})).$
%\langle p^y, as, u^z \rangle$.

\end{definition}
For an example of well defined reasons, and complete, sound and optimal explanations, see the later section ``Return to the Office at \jpmc''.%\ref{sec:exampleRTTO}.

%%%%%%%%%%%%%%%%%%%%%%%%%%%%%%%%%%%%%%%%%%%%%%%%%%%%%%%%%%%%%%%%%%%%%%%%

\section{Solving \expres Tasks using MILP}
We propose to use MILP to solve \expres problems, i.e., to compute the set of reasons that explain why a given preference was unsatisfied.
The use of MILP to solve \expres problems follows naturally, since we want to compute a set of reasons that optimise a given metric (the order of the preferences used in the reasons), subject to some constraints (well-defined reasons).
Given an \expres problem, we formulate it as a MILP as follows.\footnote{Without loss of generality, we model \expres to generate explanations for a single unsatisfied preference, but the problem can be generalised to generate explanations for all the unsatisfied preferences in a \pres problem.}

%%
%\skcomment{Please define RANK before you use it.}

\begin{equation}
\small
\label{objective_function}
\text{minimize}\quad \displaystyle\sum_{p \in \mbox{SAT}, as \in \textsc{Involved}(S,u)}x_{p,as,u} * {\cal O}(p) 
\end{equation}
subject to the following constraints:
\begin{equation}
\small
\label{all_explained}
\sum_{p \in \mbox{SAT}}  x_{p,as,u}  = 1 , \  as \in \textsc{Involved}(S,u)
\end{equation}
\begin{equation}
\small
\label{satisfied_preference_related_to_assignment}
x_{p,as,u} \leq  \textsc{Affected}(p,as) \  , \  p \in \mbox{SAT}, as \in \textsc{Involved}(S,u)
\end{equation}
\begin{equation}
\small
\label{preserved_ranking}
x_{p,as,u} \leq  \textsc{rank}(p, u)  \  , \  p \in \mbox{SAT}, as \in \textsc{Involved}(S,u)
\end{equation}
There is only one type of decision variable, $x_{p,as,u}$, which represents all the possible reasons $\langle p,as,u\rangle$ in the \expres problem: $p \in \mbox{SAT}$, $as\in \textsc{Involved}(S,u)$, and $u \in \mbox{UNSAT}$ which is the unsatisfied preference we want to explain.
The variable $x_{p,as,u}$ gets a value of $1$ if reason $\langle p,as,u\rangle$ is used in the explanation $\mathcal{E}_{S,u}$ that solves the \expres problem, and a value of 0 otherwise.
Therefore, the computational complexity of our approach depends on the number of satisfied preferences and the number of assignments to be explained returned by the \textsc{Involved} function.

Expr.~\eqref{objective_function} models the objective function of our MILP: to minimise the rank of the satisfied preferences used in the explanation's reasons.
%%
% To achieve this, the rank of the satisfied preference-type ${\cal O}(p)$ is multiplied by a big constant $\kappa$, so 
This means that the MILP tries to use more important satisfied preferences to explain unsatisfied preferences.
This objective function ensures that if the MILP finds an optimal solution, the explanation ${\cal E}_{S,u}$ extracted from such a solution is optimal (Def.~\ref{def:explanation_optimality}).

Constr.~\eqref{all_explained} ensures that we provide exactly one reason for each of the assignments returned by $\textsc{Involved}(S,u)$, i.e., for each assignment involved in $u$ being unsatisfied in solution $S$.
Therefore, if the MILP finds an optimal solution, the explanation is complete (Def.~\ref{def:explanation_completeness}).

Constr.~\eqref{satisfied_preference_related_to_assignment} ensures that we only select reasons where the given assignment is affected by the preference, $\textsc{Affected}(p,as) = 1$.
Finally, Constr.~\eqref{preserved_ranking} ensures that we only select reasons where more (or equally) important  preferences are used to justify less (or equally) important unsatisfied preferences.
These two constraints ensure that, if the MILP finds an optimal solution, the explanation is sound (Def.~\ref{def:explanation_soundness}).

In some cases, we might be interested in computing more than one explanation for a given \expres task. 
This could be the case when different users have a subjective order over the preference types that is different from the Principal's order ($\cal O$), making them prefer some reasons/explanations over others.
To compute multiple explanations, we can iteratively run the MILP, forcing the previously found solutions to not be accepted.
In this way, we get explanations with equal or higher cost (lower quality) at each iteration.

Note that this MILP formulation is general and can be used to generate explanations for any \expres task regardless of the preferences in the associated \pres problem.
Depending on the preferences in the \pres problem and its interactions, one just needs to appropriately define the \textsc{Involved} and \textsc{Affected} functions that describe when assignments are involved in or are affected by unsatisfied and satisfied preferences, respectively.
Then the MILP automatically generates the set of reasons that better explain an unsatisfied preference.

%%%%%%%%%%%%%%%%%%%%%%%%%%%%%%%%%%%%%%%%%%%%%%%%

\section{Parsing and Clustering \expres Solutions}
Explanations are the output of a cognitive process, meant to identify the necessary information to justify an event, and a social process, meant to convey that information to an explainee \cite{DBLP:journals/ai/Miller19}. So, once MILP provides us with a solution to the \expres problem, i.e., it identifies the reasons why $u$ was unsatisfied in the schedule $S$, it is crucial to make that solution understandable for any user, not only for expert ones.
In this section we present a possible approach, which we validated with users, as described later, %in section ``Evaluation through User Studies'', 
to parse the MILP explanations in natural language.
%, but alternative approaches could be defined. 
Conveying MILP solutions can be challenging, especially (i) when the $\textsc{Involved}$ function returns many assignments that need to be explained; and (ii) because the variables in the MILP solution contain information about the agents and their preferences: to include them in an explanation may be beneficial in some cases, but harmful in others where the privacy of the agents needs to be preserved~\cite{kraus2020ai}.

% One challenging aspect of conveying MILPs solutions is that these can be potentially long (i.e., containing many reasons) if the $\textsc{Involved}$ function returns many assignments that need to be explained. 
% Another complication arises from the variables in the MILP solution containing information about the agents and their preferences: to include it in an explanation may be beneficial in some cases, but harmful in others where we want to preserve the privacy of the agents~\cite{kraus2020ai}.

According to the application context, we recommend the definition of natural language templates describing constraints, preferences and reasons. 
%For example, a constraint $c \in \mathcal{C} : R \times T \longrightarrow \mathbb{N}$ can be expressed as ``there are $n$ [resources] available on $t$'', where $n,t$ and resources are variables appropriately instantiated.
% Then, we consider that preferences can be combined to describe a reason and reasons can be combined in order to provide an explanation.
%%
% We parse an assignment $as=\langle ag, r, t\rangle$ with the following template:
% \begin{quote}
%     ``$as(ag)$ was assigned $as(r)$ on $as(t)$"
% \end{quote}
% Likewise, w
For instance, a reason ${\cal R}=\langle p, as, u \rangle$ can be parsed with the following template: \textit{``$ag(p)$ was assigned $r(as)$ on $t(as)$ instead of $ag(u)$ because to satisfy $\tau(p)$ is more important than to satisfy $\tau(u)$"}.
As we mentioned, the solution returned by the MILP contains very granular information (for each assignment), even after parsing it in natural language.
This can be useful for Principals, as it allows them to understand all the details, but might yield explanations that are tedious and difficult to interpret for general users.
%
%As we mentioned, according to the MILP solution, the list of reasons to be included in the explanation may be long and contain too much information.
%repetitive, for instance if all the reasons are due to satisfied preferences of the same type. 
For this reason, we suggest to aggregate the reasons according to time slot and preference type before parsing: e.g., if there are $n$ reasons $\langle p, as, u \rangle$ justifying the assignments of $n$ agents having satisfied preferences of the same type on the time slot $t$, we could have ``$ag(p_1)$, ..., $ag(p_n)$ were assigned on $t(as)$ instead of $ag(u)$ because to satisfy $\tau(p)$ is more important than to satisfy $\tau(u)$''.

Finally, given an explanation $\mathcal{E}_{S,u}$ of why the preference $u$ was not satisfied by the solution $S$ to the \pres problem, we parse it as follows:
% \begin{quote}
    \textit{``$u$ could not be satisfied because 
    [list of constraints $\cal C$] 
    [list of reasons in $\mathcal{E}_{S,u}$]''}.
% \end{quote}
Lastly and if the scenario requires it, we recommend to remove any identifying reference to the agents whose satisfied preferences are mentioned in the explanation, in order to better preserve their privacy.
For a complete example of parsed explanations, see the next section, where we discuss an \expres problem and solution in the context of workforce scheduling, an application of interest at \jpmc.

%%%%%%%%%%%%%%%%%%%%%%%%%%%%%%%%%%%%%%%%%%%%%%%%%%%%%%%%%%%%%%%%%%%%%%%%

\section{Return to the Office at \jpmc}\label{sec:exampleRTTO}
\begin{figure}
    \centering
    \includegraphics[width=\columnwidth]{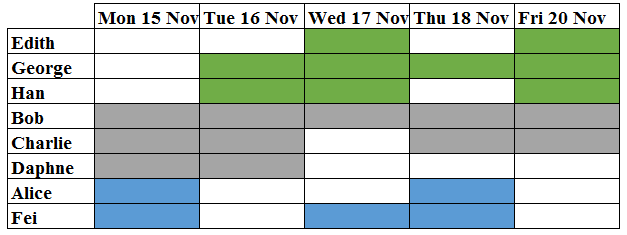}
    \caption{Solution to the \pres problem in the working example.}
    \label{fig:example-schedule}
\end{figure}

In this section we exemplify the \expres problem formulation and solution through MILP that we have thus far introduced in a general manner. In order to do this, we refer to a real-world scenario, namely \textit{return to the office} at \jpmc.
As outlined in the introduction, in this \pres problem a set of employees ($Ag$) needs to be assigned to a limited number of desks $R$ (that is the only resource type) over a fixed time period ($T$). In the example we discuss, summarised by Tab.~\ref{tab:example-preferences} and Fig.~\ref{fig:example-schedule}, we consider 8 employees and a time period of one working week (the 15th to the 20th of November, 2021). Coloured cells mark the assignments of employees to the office, with different colours representing different working groups (a set of agents that need to be co-assigned).
The company, which acts as Principal, defines the following set of constraints $\cal C$ (in the depicted example, $n_{desks}=5$).
For simplicity, here we assume assignments to be binary variables, i.e., $y_{ag, r, t} = 1$ if $\langle ag, r, t \rangle \in S$, and $0$ otherwise.
%The only constraint of the problem is that for each day ($t \in T$), the sum of employees assigned to a desk must be equal to the number of available desks defined by the Principal:
\begin{equation}
\small
    \label{max_desks}
    \sum_{ag \in Ag}  y_{ag,r,t} = n_{desks}, \ r \in R, t \in T
\end{equation}
\begin{equation}
\small
    \label{max_days}
    \sum_{t \in T} y_{ag, r, t} \leq \mbox{maxDays}(ag), \ r \in R, ag \in Ag
\end{equation}
\begin{equation}
\small
    \label{day_out}
    y_{ag, r, t} =  0  \quad \text{ if }  \mbox{dayOut}(ag,t), \ ag \in Ag, r \in R, t \in T
\end{equation}
Constr.~\eqref{max_desks} ensures that the sum of the employees assigned to a desk on the same day is equal to the number of available desks defined by the Principal.
%%
%\skcomment{How the constraints follow the definition A \textbf{constraint} $c \in \mathcal{C}$ is a function that maps any combination of desirable elements ${\cal D}$ onto $\mathbb{N}$. }
Constr.~\eqref{max_days} ensures that employees are not assigned to the office more days than the maximum they requested. 
Constr.~\eqref{day_out} ensures that employees are not assigned on the days they are out of office, for example due to vacations or personal matters.

\begin{table}[]
    \centering
    \scriptsize
    \begin{tabular}{c|c|c|c|c|c}
    \toprule
        Employee & Min-Max Days & Preferred & Group & Meetings & OOO \\
        \midrule
        Edith & 1 - 2 & Th & 1 & 17/11&\\
        George & 4 - 5 & & 1 && 15/11\\
        Han & 3 - 4 & W,Th,F & 1 & 17/11 &\\
        Bob & 5 - 5 & & 2 & 18/11 &\\
        Charlie & 4 - 4 & F & 2 && 17/11\\
        Daphne & 2 - 3 & M,Tu,Th & 2 & 15/11 &\\
        Alice & 2 - 4 & M,Tu & 3 & 18/11 &\\
        Fei & 3 - 4 & W,Th,F & 3 & 15/11,17/11&\\
         \bottomrule
    \end{tabular}
    \caption{Scheduling preferences of the employees in the working example.}
    \label{tab:example-preferences}
\end{table}

Regarding the employees' preferences, we consider the following totally ordered set of preferences types ${\cal O} = \langle \tau_{min} \prec \tau_{meet} \prec \tau_{group} \prec \tau_{pref} \rangle$, as imposed by the Principal in \jpmc, where: 
\begin{itemize}
    \item $\tau_{min}$ represents a preference type where an employee asks to have a desk at least $n \in \mathbb{N}$ days over the time period $T$ (instantiated as $p =\langle min, ag, n \rangle$).
    
    \item $\tau_{meet}$ represents a preference type where an employee asks to have a desk on a given day because he/she has an important meeting (instantiated as $p = \langle meet, ag, t \rangle$). 
    
    \item $\tau_{group}$ represents a preference type where an employee asks to have a desk together with another employee on a given day because they need to collaborate (instantiated as $p = \langle group, ag_1, ag_2, t \rangle$). 
    
    \item $\tau_{pref}$ represents a preference type where an employee asks to have a desk on a given weekday, e.g., for personal convenience (instantiated as $p = \langle pref, ag, t \rangle$). 
\end{itemize}

We can explain any unsatisfied preference, but in our example we focus on
% solve through MILP any \expres problem, but we focus, as an example, on 
explaining $\widehat{u} = \langle pref, Edith, Thursday \rangle$, i.e., why Edith's preferred day on Thursday could not be satisfied.

% In order to identify the assignments in the schedule that are involved with or affected by preferences, 
We instantiate the \textsc{Involved} and \textsc{Affected} functions as follows. 
$\textsc{Involved}(S,u)$ returns the set of assignments $S^\prime \subseteq S$ involved in an unsatisfied preference $u$ depending on its type:
\begin{itemize}
    \item If $\tau(u) = min$, \textsc{Involved} returns all of the assignments in $as \in S$ where $ag(u) \neq ag(as) $, in order to provide a reason why the rest of the desks were better assigned in that way and could not be assigned to $ag(u)$.
    
    \item If $u$ is of any other type, \textsc{Involved} returns all of the assignments in $S$ where $t(as) = t(u)$, in order to provide a reason why each of the desks on day $t$ were better assigned to other agents and could not be assigned to $ag(u)$.
\end{itemize}
For example, when considering $\widehat{u}$, %=\langle \mbox{Edith, Thursday} \rangle$,
$\textsc{Involved}(S,\widehat{u})$ returns the five assignments of agents to desks on Thursday: $\{\langle \mbox{Bob, Thursday} \rangle, \langle \mbox{Charlie, Thursday} \rangle,\ldots\}$.

Likewise, $\textsc{Affected}(p, as)$ checks if a satisfied preference $p$ is related to an assignment $as$ depending on the preference type:
\begin{itemize}
    \item If $\tau(p) = min$, \textsc{Affected} returns $1$ if $ag(p)=ag(as)$ and $|T| - \sum_{t \in T}\mbox{dayOut}(ag(p),t) = n(p)$. That is, a satisfied preference of this type affects an assignment if the number of days that the employee is available over the time period is equal to his/her minimum, i.e., his/her days at the office cannot be reduced; and if the employee in the preference is the same as the employee in the assignment.
    
    \item If $\tau(p) = meet$ or $\tau(p) = pref$, \textsc{Affected} returns $1$ if $ag(p)=ag(as)$ and $t(p)=t(as)$. That is, a satisfied preference of these types affects an assignment if the employee was assigned a desk on the day he/she requested. 
    
    \item If $\tau(p) = group$, \textsc{Affected} returns $1$ if $ag_1(p)=ag(as)$, $t(p)=t(as)$, and $\exists as_2 \in S \mid ag_2(p)=ag(as_2)$, $t(p)=t(as_2)$. That is, a satisfied preference of this type affects an assignment if both employees are assigned a desk on the same day.
\end{itemize}
% In our running example, considering a satisfied preference $p=\langle\mbox{group, Edith, George, Wednesday}\rangle$, $\textsc{Affected}(p,as)$ would return $1$ to $as = \langle \mbox{Edith, Wednesday} \rangle$, since the preference is related to the assignment, and $0$ to $as = \langle \mbox{Han, Tuesday}\rangle$ since the preference is not related to that assignment.
In the running example, considering a satisfied preference $p'=\langle group, Edith, George, Wednesday\rangle$, we have $\textsc{Affected}(p',\langle Edith, desk, Wednesday \rangle)=1$ because the preference is related to the assignment; and $\textsc{Affected}(p',\langle Han, desk, Tuesday \rangle)=0$ because the preference is not related to that assignment.

When explaining why $\widehat{u}$ was unsatisfied in the schedule $S^*$, 
% an explanation is \textit{complete} iff it is able to provide a reason for all the involved assignments, with every reason using a different agent in the \francesca{satisfied} preference side; an explanation is \textit{sound} iff all the reasons within it are well-defined.
% For instance,
to say that Edith could not be assigned on Thursday because Daphne was assigned on Monday is an ill-defined reason, but to refer to Bob being assigned on Thursday due to a meeting is a \textit{well-defined} one (cf. Def.~\ref{def:well_defined_reason}). 
An example of a \textit{complete}, \textit{sound} and \textit{optimal} explanation (cf. Defs.~\ref{def:explanation_completeness},\ref{def:explanation_soundness},\ref{def:explanation_optimality}) is \textbf{E}: ``The preference could not be satisfied because the 5 available desks were assigned to other people with more important preferences: George, Bob and Charlie due to a minimum number of days per week; Alice due to meetings; and Fei due to 1 working group.''
\expres can generate other explanations, not optimal but still sound and complete, that may be of interest according to the circumstances. For example, Alice's assignment could be justified due to her working group, without mentioning her meeting as in \textbf{E}, in case that meeting was confidential, or if Edith considers more convincing explanations regarding working groups (if her subjective order over the preference types is different from the Principal's one).

\section{Evaluation through Simulation}
We evaluate our approach by providing explanations in simulated scenarios of our return to the office domain.
\subsection{Experimental Setting}
We generate problems in three configurations of increasing size: 10, 30 and 50 employees over a fixed period of a five-day week.
For each configuration, we generate 100 \pres problems with random preferences for each agent.
Each agent randomly has 1 or 2 meetings, 1 or 2 preferred days and 1 to 4 working group preferences. The number of days on which an agent has a preference defines his/her preference for minimum number of days; as a maximum number of days, they have a random number between their minimum and $5$.
%%
% As the minimum number of days they have the number of days they have a preference on; as maximum number of days, they have a random number between their minimum and $5$.
%%
For example, if an agent has preferences over $2$ different days, its minimum number of days is $2$ and its maximum is a random number between $2$ and $5$.
Agents also have a $20\%$ probability of having dates out (1 or 2, randomly picked).
%%
% We make sure the problems are over constrained, i.e., no solution can accommodate all employees' preferences, by setting
We set the number of available desks each day to be $50\%$ of the total number of employees, in line with the policy followed in the actual return to the office at \jpmc.

We optimally solve all of these \pres problems and automatically compute all of the satisfied and unsatisfied preferences.
We have an average of $31.4$ satisfied and $28.7$ unsatisfied preferences in the problems with $10$ employees; $89.4$ and $141.7$ in the problems with $30$ employees; and $147.5$ and $280.1$ in the problems with $50$ employees.
For each problem, we randomly pick one unsatisfied preference of each type (if one exists) to build the \expres task to be solved by the MILP.
This gives us a total of $1200$ ($3$ agent configurations, $100$ problems on each configuration, and $4$ unsatisfied preferences on each problem) \expres tasks to solve.
We run the MILP on each task with a timeout of $30$ seconds, or after $1000$ explanations are produced.
We use CPLEX\footnote{https://www.ibm.com/analytics/cplex-optimizer} to solve the MILP.
Experiments were run in Intel(R) Xeon(R) CPU E3-1585L v5 @ 3.00GHz machines with 64GB of RAM.

\begin{figure*}
    \centering
    \includegraphics[width=.32\textwidth]{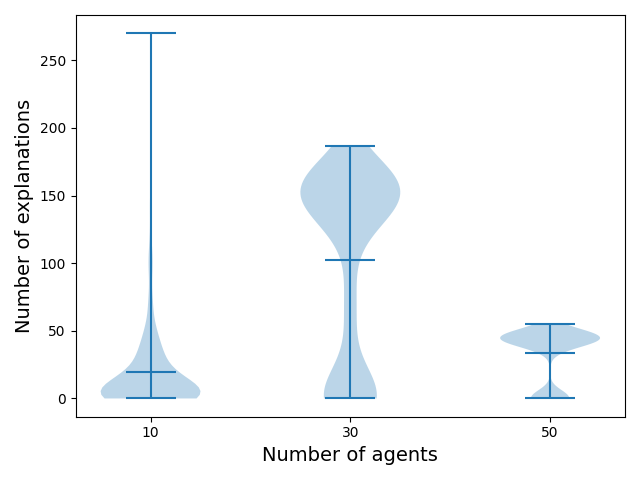}\hfill
\includegraphics[width=.32\textwidth]{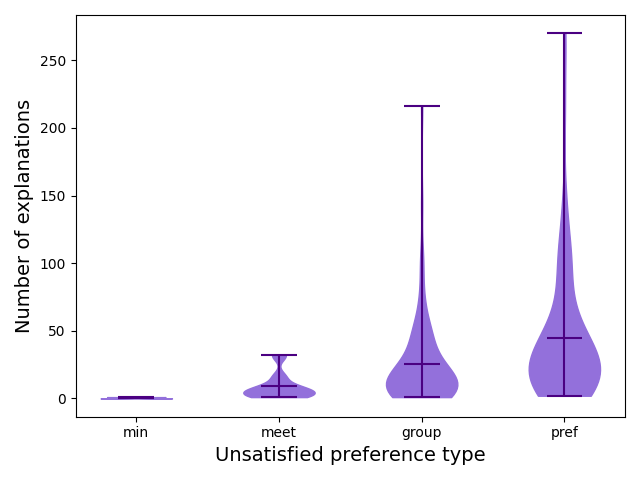}\hfill
\includegraphics[width=.32\textwidth]{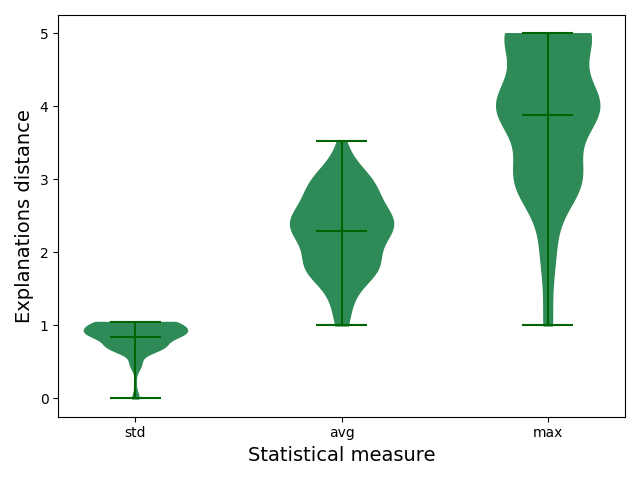}
    \caption{The figure on the left shows the number of explanations generated within 30 seconds as we increase the number of agents. The figure in the middle shows the number of explanations we can compute depending on the unsatisfied preference type in problems with $10$ agents. The figure on the right shows different statistical measures over the distance between the generated explanations.
    For each violin plot, the central horizontal line depicts the mean of the distribution, while the other two lines depict the minimum and maximum values. A wider shadow indicates that more points have that value.}
    \label{fig:simulation_experiments}
\end{figure*}

\subsection{Scalability Evaluation}
% \francesca{the scalability conclusions were confusing for the reviewers: maybe we can first say that we were always able to compute the optimal explanation in a very short time; then, we talk about generating all the sounds explanations, for which we timeout.}
First, we evaluate the scalability of our approach, by measuring the time needed to compute the first (optimal) explanation and the number of explanations provided for each scenario (see left plot of Figure~\ref{fig:simulation_experiments}). 
\expres tasks can be solved in a reasonable time for all the configurations: even with 50 agents, the solver returns the optimal explanation on average in less that 0.5 seconds.
Regarding the number of generated explanations, we expect it to increase as we increase the number of agents, because with more agents there are more satisfied preferences and more ways of combining them to justify unsatisfied preferences. However, when increasing the number of agents, the complexity of the problems also increases, allowing less problems to be solved within the time bound.
%%
% Left plot of Figure~\ref{fig:simulation_experiments} shows the number of explanations generated as we increase the number of agents.
%%
With $10$ agents, we are able to generate all of the sound explanations in less than $1$ second.
%, and get the first explanation always in less than a second.
%%
For these problems, we can compute an average of $19.7$ different explanations, with some \expres tasks where we can compute more than $250$ explanations.
With $30$ and $50$ agents we cannot generate all of the sound explanations within the given time bound. Without timing out, the solver generates an average of $157.2$ and $402.3$ different explanations with 30 and 50 agents, respectively.  
We conclude that \expres is scalable because, even though not \textit{all} the sound explanations are found, the user will ultimately be interested in only one: how to identify and learn the preferred explanation for a user is part of our future work.

% On the other hand, the number of explanations tends to drop with $50$ agents.
%%
% However, with 50 agents we generated less explanations due to the given time bound
% This occurs because these problems are harder and the solver needs an average of $0.5$ seconds to return the first (optimal) explanation, thus generating less explanations in the allowed time than in problems with $30$ agents.
%%

\subsection{Explanations per Unsatisfied Preference Type}
Next, we analyse how the number of explanations is influenced by the type of unsatisfied preference that needs to be explained (see the central plot of Figure~\ref{fig:simulation_experiments}). We consider only the problems with $10$ agents given that it is the only configuration for which we can compute all of the explanations in all of the problems.
As expected, we can compute more explanations if we explain less important, unsatisfied preferences such as those of type $\tau_{pref}$, where we can produce an average of $44.5$ explanations for each unsatisfied preference.
This happens because there is a larger number of more important preferences that can be used to explain these unsatisfied preferences.
Coherently, there are fewer ways of explaining more important preferences such as preferences of type $\tau_{meet}$ being unsatisfied, since they can only be explained by using preferences of type $\tau_{meet}$ or $\tau_{min}$, that are equally or more important.
In fact, there are approximately $35\%$ of problems involving an unsatisfied preference in $\tau_{min}$ where we cannot produce any explanation.
This is because these problems were extremely overconstrained and the solution could not satisfy many requests for a minimum number of days.
Since our reasons require satisfied preferences, there were some cases where we could not produce any explanation.

\subsection{Understanding Explanation Structure}
Finally, given an \expres problem and the set of generated explanations, we investigate how different the explanations within that set are.
We focus on the 100 problems with $10$ agents where the selected unsatisfied preference is of type $\tau_{pref}$, since these are the cases where we can generate more explanations.
Then, given two explanations ${\cal E}_1$ and ${\cal E}_2$, we measure their distance as $|{\cal E}_1 \setminus {\cal E}_2|$, i.e., the number of different reasons they contain.
Given the 5 available desks, the distance is bounded between $0$, if both explanations are the same, and $5$, if the two explanations have no common reason.
We compute this pairwise distance for all of the pairs of explanations produced in an \expres problem (see the right-hand plot of Figure~\ref{fig:simulation_experiments}).
%%
% We do this for the $100$ problems that fall on that category and report the results in the right plot of Figure~\ref{fig:simulation_experiments}.
%%
As we can see, the standard deviation of the explanations' pairwise distance in each problem is close to $1$, meaning that explanations tend to vary for one out of five reasons.
The average distance between explanations in a problem is $2.3$, and the average maximum distance is $3.8$, with more than $75\%$ of the problems having a maximum distance between explanations higher than $2.7$.
These results suggest that many of the explanations we provide for a problem only differ in $1-2$ reasons ($20-40\%$ of the explanation).
However, most of the time our set of explanations contains at least two explanations that are really different, differing in $3-4$ reasons ($60-80\%$ of the explanation).

%Finally, we take a deeper look to the explanations we produce for each problem by analyzing the reasons appearing on them.
%%
%To do this we use the \expres problem we already discussed in the previous Return to the Office at \jpmc section, which we will also employ in our user studies.
%%
%Right plot of Figure~\ref{fig:simulation_experiments} shows the different cost of the $24$ explanations we can produce in that scenario.
%%
%Each increase in the cost function corresponds to lower quality reasons entering in that explanation with respect to the previous one.
%%
%Likewise, plateaus in the cost function corresponds to different explanations that employ the same number of preferences types.
%%
%There is only one optimal explanation that uses $3$ satisfied preferences related to minimum number of days, $1$ to a meeting, and $1$ to a working group to justify why Edith preference over Thursday could not be satisfied.
%%
%The big cost increase after explanation $12$ corresponds to the introduction of reasons with equally important preferences in the explanations.
%%
%The worst explanation we can produce in this scenario uses $1$ minimum number of days, $3$ working groups and $1$ preferred day (equally important preference) to justify Edith's unsatisfied preference.

%\alberto{AP: if we want to enrich the analysis we can also show how the number of explanations is affected by the number of preferences: a lot of preferences $\rightarrow$ more explanations. However I think we should keep the evaluation through simulations short so we can spend more time explaining the user studies.}

\section{Evaluation through User Studies}
We have designed and implemented two user studies in order to understand (i) how humans solve \expres tasks, (ii) how automated generated explanations compare to human generated ones, and (iii) what type of explanations are preferred by users of the tool. 
In particular, we wanted to validate the following hypotheses:
\begin{itemize}
    \item \textbf{Hp1}: The \expres framework produces explanations faster than humans.
    % \item \textbf{Hp2}: Human-generated explanations are less sound, complete and optimal than automatically-generated ones.
    \item \textbf{Hp2}: Humans find automatically generated explanations at least as satisfying as the human generated ones.
    % \item \francesca{Different people prefer explanations including different features.}
\end{itemize}
The first user study (US1), where we collected human-generated explanations, aimed to discuss Hp1; the second user study (US2), where we compared human-generated and \expres-generated explanations, aimed to discuss Hp2.
We defined two \pres scenarios with different complexities by considering 8 employees in the first scenario (see Tab.~\ref{tab:example-preferences} and Fig.~\ref{fig:example-schedule}) and 20 employees in the second scenario (see Supplementary Material), to be scheduled over a week. In US1 we showed all the preferences and the entire team's weekly schedule; in US2 we showed only one individual set of preferences and one individual weekly schedule.

\subsection{User study 1}
US1 consists of individual interviews with people (N=10) who have previously actively interacted with the workforce scheduling tool deployed at \jpmc, i.e., team managers or assistants who generated schedules through the tool. The interviews took place virtually. After enquiring about the participant's familiarity with the tool (on a 5-point Likert scale, avg=4.9), the two scenarios were fully disclosed. The participants were asked to justify to one selected fictitious employee why their preference could not be satisfied in the team's schedule (e.g., ``Why wasn’t Edith assigned to the office on Thursday 18th November, as she requested?''). In the meeting chat, the participants wrote
%\footnote{Note that, because of company policies, it was impossible to record the interview.} 
an explanation for the unsatisfied preference and then evaluated the difficulty of providing that explanation. We tracked the time that each participant required to provide each an explanation.

\begin{table}[]
\centering
\scriptsize
\begin{tabular}{l|lll|lll}
\toprule
           & \multicolumn{3}{c|}{Scenario 1} & \multicolumn{3}{c}{Scenario 2} \\
           & avg          & std dev & \expres       & avg           & std dev  & \expres      \\
                   \midrule
time (s)       & 195.9        & 80.8   & $\ll$ 1        & 186.3         & 96.2  & 0.8          \\
difficulty & 2.4          & 1.4    & --        & 2.7           & 1.1 & -- \\
\bottomrule
\end{tabular}
\caption{Results of user study 1. Time is expressed in seconds, difficulty is expressed on a 5-point Likert scale (1=very easy, 5=very difficult).}
\label{tab:us1-results}
\end{table}

Tab.~\ref{tab:us1-results} shows a quantitative overview of the results. All of the participants spent considerably more time than \expres to provide an explanation (Hp1 is confirmed): despite scenario 2 being more complex in terms of quantity and density of information and being reported more challenging to explain (avg difficulty=2.7 vs 2.4 in scenario 1), 6 participants were quicker to provide the second explanation than the first one. We suppose this could be due to the participants being more familiar with the task by the time they faced the second scenario. Regarding the quality of explanations, in each scenario we gathered very different justifications (see Tab.~\ref{tab:human-explanations} for a sample, and Supplementary Material for a complete list of the collected explanations), sometimes more explicit and detailed% (s1,p2: ``Edith wasn't assigned into the office because she hadnt requesred the dates in on the 18th or because since her minimum was 1 and max was 2 and it maxed out the days she can come in'')
, sometimes more implicit and general% (s1,p5: ``Too many of the team members have requested to be in the office on a Thursday.'')
. However, no explanation can be considered complete, sound or optimal, according to Defs.~\ref{def:explanation_completeness}, \ref{def:explanation_soundness} and \ref{def:explanation_optimality}.

\begin{table}[]
    \centering
    \scriptsize
    \begin{tabular}{p{80mm}}
    \toprule
    \textbf{Scenario 1} \\
    \textbf{H4:} Too many of the team members have requested to be in the office on a Thursday.\\
    \textbf{H5:} Edith wasn't assigned into the office because she hadnt requesred the dates in on the 18th or because since her minimum was 1 and max was 2 and it maxed out the days she can come in\\
    \textbf{H6:} 4 desks on Thursday accounted for 2 people had in office dates \& 2 people had min4 days in and out of office dates on another day meaning that they would be in all other day of the week.   \\
        \midrule
    \textbf{Scenario 2} \\
    \textbf{H4:} Due to being a large group and many of the employees requesting the same day, not everyone could get their desired day in the office.\\
    \textbf{H5:} 3 people had dates in, another 2 had 5 days min, the rest are perhaps a combination of all the other factors, meaning they had to be in instead of Ivan\\
    \textbf{H6:} He was limited by the seat availability that day , members of two other groups has selected the 1th specifically to be in so they took preference \\
    \bottomrule
    \end{tabular}
    \caption{Details of the human-generated explanations selected from US1 and included in US2.}
    \label{tab:human-explanations}
    \vspace{-10pt}
\end{table}

\subsection{User study 2}
US2 consists of an online questionnaire with people (N=28) who have passively interacted with the tool, i.e., employees whose schedule was generated by the tool, within \jpmc. Each participant was shown, sequentially, the preferences and the schedule of one fictitious employee (e.g., Edith's) from the two scenarios previously defined. For each scenario, 6 explanations were listed in a random order: 3 were generated by \expres (E1-E3, see Tab.~\ref{tab:explanations-US2}; explanations are anonymised and parsed as shown at the end of ``Return to the Office at \jpmc'') and 3 were selected from the human generated ones in US1 (H4-H6, see Tab.~\ref{tab:human-explanations}). When selecting the explanations to include in this study, we aimed to represent the diversity, in terms of structure and reasons mentioned, of the pool of explanations that were available, in order to explore the participants' appreciation of different combinations of reasons. In both scenarios, E1 is the optimal explanation (cf. Def.~\ref{def:explanation_optimality}). 
Each participant was asked to first select and then rank the three explanations that were the most satisfying. We report the results in Tab.~\ref{tab:resultsUS2}.

\begin{table}[]
    \centering
    \scriptsize
\begin{tabular}{@{}l|ccc|ccc@{}}
\toprule
\multirow{2}{*}{Reason type} & \multicolumn{3}{c|}{Scenario 1 ($n$=5 desks)} & \multicolumn{3}{c}{Scenario 2 ($n$=12 desks)} \\
                             & E1   & E2   & E3   & E1   & E2   & E3   \\ \midrule
$R_{min}$                      & 3        & 1        & 1        & 3        & 0        & 3        \\
$R_{meet} $                    & 1        & 0        & 2        & 2        & 0        & 0        \\
$R_{group}$                    & 1        & 4        & 1        & 7        & 12       & 8        \\
$R_{pref} $                    & 0        & 0        & 1        & 0        & 0        & 1     \\ \bottomrule  
\end{tabular}
    \caption{Details of the reasons part of the explanations generated by \expres and included in US2. \textit{Reason type} refers to the preference-type mentioned as a justification for the unsatisfied preference.
    % As an example, S1-E1 is parsed and anonymised as: ``The preference could not be satisfied because the 5 available desks were assigned to other people with more important preferences: 3 employees due to minimum number of days per week; 1 employee due to meetings; 1 employee due to 1 working group.''
    }
    \label{tab:explanations-US2}
\end{table}

Participants showed a strong preference in both scenarios for the \expres explanations, which were selected significantly (t-test with pvalue=0.01) more often than the human-generated ones (Hp2 is confirmed).
% To better understand the participants' behaviour, we checked whether their selection of explanations was consistent across the two scenarios. To do this, we first defined $D_i$ as the distribution of the number of explanations of a given type (e.g., generated by \expres) selected by each participant in the scenario $i$; then, we investigated whether $D_1$ and $D_2$ were equivalent ($H_0: D_1 = D_2$).
% % , in order to understand the consistency of the participants' selection across scenarios. 
% Note that the mean and the standard deviation of the distribution $D_1-D_2$ are 0.14 and 0.79 respectively.
% A Kolmogorov-Smirnov test (statistic=0.11, pvalue=0.99) does not allow to reject the null hypothesis $H_0$ and suggests that the participants behaved consistently when facing the two scenarios. 
Looking at the results in greater detail, we see that E1 has been the most selected explanation in both scenarios (by 78.6\% and 82.1\% of the participants). We interpret this as a predilection for explanations that are sound, complete and as consistent as possible with the Principal's total order of preferences. Regarding the human-generated explanations, most people appreciated H4, which in both scenarios was the most general and vague explanation. This suggests that brief and simple explanations can satisfy the general user who does not look for detailed justifications.

% otherwise:
% ($H_0: \mu_{D_1} = \mu_{D_2}$). Given the result of a t-test (statistic=0.679, pvalue=0.5), we cannot reject the null hypothesis and we conclude that the two distributions are not significantly different.

\begin{table}[]
\centering
\scriptsize
\begin{tabular}{ll|llll|llll}
\toprule
 &       & E1 & E2 & E3 & \textbf{tot.E} & H4 & H5 & H6 & \textbf{tot.H} \\ \midrule
\multirow{2}{*}{s1} & \# selections & 22      & 19      & 14      & \textbf{55}          & 17     & 5      & 7      & \textbf{29}        \\
                            & rank score     & 47      & 36      & 24      & \textbf{107}         & 37     & 10      & 14     & \textbf{61}        \\ \midrule
\multirow{2}{*}{s2} & \# selections & 23      & 17      & 11      & \textbf{51}          & 14     & 8      & 11     & \textbf{33}        \\
                            & rank score     & 55      & 28      & 17      & \textbf{100}          & 34     & 12     & 22     & \textbf{68}        \\ \bottomrule

\end{tabular}
\caption{Results of US2. The rank score is $3x_1+2x_2+x_3$ ($x_i$ is the n. of times an explanation has been ranked $i$).}
\label{tab:resultsUS2}
\vspace{-5pt}
\end{table}

%%%%%%%%%%%%%%%%%%%%%%%%%%%%%%%%%%%%%%%%%%%%%%%%%%%%%%%%%%%%%%%%%%%%%%%%

\section{Related Work}
Explanations are essential for humans to understand the outputs and decisions made by AI systems~\cite{DBLP:conf/aaai/CoreLLGSR06,DBLP:journals/ai/Miller19}.
There exist many works that provide explanations for different AI use cases ranging from automated planning~\cite{DBLP:journals/corr/abs-1709-10256,DBLP:conf/aips/ChakrabortiKSSK19} to machine learning~\cite{carvalho2019machine} or deep learning~\cite{samek2017explainable}.
Explanations are also crucial in multi-agent environments where some extra challenges arise, such as privacy preservation or fairness~\cite{kraus2020ai}.
Scheduling of multiple agents is one of these problems, and explaining the resulting schedules is not a trivial task.

%%
% Other works have also focused on providing explanations to AI generated schedules.

In \cite{DBLP:journals/corr/abs-2011-08733}, the authors propose \textsc{crosscheck}, a tool that (i) diagnoses scheduling failures in the context of a Mars Rover mission, and (ii) guides users about which constraints need to be altered in order for the activity to be successfully scheduled.
% The tool was evaluated positively by intended users, who found its explanations helpful to adjust the schedules as needed. 
Their tool focuses on visualisation and improving user experience with the scheduler, but could hardly be adapted to provide explanations for multiple users having competing preferences.
% solve \expres problems, where multiple agents having multiple preferences 

% In \cite{DBLP:journals/corr/abs-2011-08733}, the authors propose \textsc{crosscheck}, a tool that explains why activities failed to be scheduled in the context of a Mars Rovers mission. 
% %%
% In their case, they provide guidance on potential constraint relaxations that would enable the schedule of these activities, while we try to explain why some preferences could not be satisfied.
% %%
% They show examples of how the tool is used and briefly report some feedback they got from intended users of \textsc{crosscheck}.
% %%
% In our case, we conducted a user study with actual users of the tool within \jpmc.

In \cite{DBLP:journals/corr/abs-2002-01640}, the authors propose \textsc{aita}, a centralised Artificial Intelligence Task Allocation that simulates a negotiation between the agents.
If unhappy with the recommended allocation, an agent can question it using counterfactuals; these will be refuted by \textsc{aita}, which explains, using negotiation trees, how the agent's proposal would entail a worse-off allocation than the recommended one.
Despite not formally allowing counterfactual queries, in this paper we still enable users to get explanations (i) that are specifically targeted for preferences that are unsatified in the recommended schedule, and (ii) that contain more interpretable information (\textsc{aita} refers to the overall allocation cost, while we include other agents' satisfied preferences).
Finally, the authors discuss the length of the explanations when 2-4 agents are involved in the scheduling process but do not share any results on the 
%%
% Like us, they also assume a centralised decision maker.
%%
% However, they do not focus on preserving the privacy of the agents and their preferences.
%%
% On the validation side, they also evaluate their approach with users, but they provide little evidence on the
scalability of their approach (more agents and more tasks).

In \cite{DBLP:conf/aaai/CyrasLMT19}, the authors explain schedules using argumentation frameworks. 
They can explain (i) why a solution is not feasible or suboptimal (we assume feasible and optimal solutions are given); or (ii) why a preference was not satisfied in the solution, as in our \expres problem formulation.
In order to provide the explanations, they need to manually generate the attack graphs, i.e., the relationship between the preferences and the assignments.
This is similar to the effort needed to define the rules inside our \textsc{Involved} and \textsc{Affected} functions.
A key difference between both works is that they are restricted to makespan scheduling problems with a very limited number of preferences.
Our \expres framework can be used to generate explanations in any scheduling problem where there is a totally ordered set of preferences.
On the evaluation side, they do not report any experiment. %\alberto{Even though I think they have another paper in which they talk about a tool, need to double check} 
An interactive tool was presented \cite{vcyras2021schedule}, but its empirical validation was left for future work.

%%%%%%%%%%%%%%%%%%%%%%%%%%%%%%%%%%%%%%%%%%%%%%%%%%%%%%%%%%%%%%%%%%%%%%%%

\section{Conclusion and Future Work}
In this paper, we introduced the \expres framework, an approach to explain why a given preference was unsatisfied in a schedule.
We framed this problem as an optimisation task that aims to find the best set of reasons that can explain an unsatisfied preference, and we solved it using MILP techniques.
Then we showed how to group and translate the raw explanations in order to be easily interpreted by humans as well as to preserve agents' privacy.
Experimental results through simulations showed that \expres can efficiently scale.
Finally, a set of user studies within \jpmc showed how employees interacting with a workforce scheduling tool preferred our automatically-generated explanations over human-generated ones.

Currently, we assume a totally ordered set of preferences that is always respected in any optimal solution.
This simplifies the definition of the \textsc{Involved} and \textsc{Affected} functions, but that assumption does not hold in all scheduling problems.
%%
% We will explore how to define both functions in problems where such total order does not exist, and how to define reasons that justify unsatisfied preferences through a chain of satisfied ones, without limiting to only one. 
We will explore how to treat problems where only partial orders over the preferences exist, and how to define reasons that justify unsatisfied preferences through a chain of satisfied ones, without limiting to only one. 
Also, we would like to investigate (i) whether providing explanations improves the users' satisfaction with their schedules \cite{bradley2009dealing}, and (ii) how to learn the subjective preference order of a user and the type of explanations they prefer, in order to generate even more tailored explanations~\cite{DBLP:journals/corr/abs-2106-12207}.

\section*{Acknowledgements}
This paper was prepared for informational purposes in part by
the Artificial Intelligence Research group of JPMorgan Chase \& Co. and its affiliates (``JP Morgan''),
and is not a product of the Research Department of JP Morgan.
JP Morgan makes no representation and warranty whatsoever and disclaims all liability,
for the completeness, accuracy or reliability of the information contained herein.
This document is not intended as investment research or investment advice, or a recommendation,
offer or solicitation for the purchase or sale of any security, financial instrument, financial product or service,
or to be used in any way for evaluating the merits of participating in any transaction,
and shall not constitute a solicitation under any jurisdiction or to any person,
if such solicitation under such jurisdiction or to such person would be unlawful.

%%%%%%%%%%%%%%%%%%%%%%%%%%%%%%%%%%%%%%%%%%%%%%%%%%%%%%%%%%%%%%%%%%%%%%%%

%%% The acknowledgments section is defined using the "acks" environment
%%% (rather than an unnumbered section). The use of this environment 
%%% ensures the proper identification of the section in the article 
%%% metadata as well as the consistent spelling of the heading.

%%%%%%%%%%%%%%%%%%%%%%%%%%%%%%%%%%%%%%%%%%%%%%%%%%%%%%%%%%%%%%%%%%%%%%%%

%%% The next two lines define, first, the bibliography style to be 
%%% applied, and, second, the bibliography file to be used.

\bibliographystyle{aaai} 
\bibliography{ref}

%%%%%%%%%%%%%%%%%%%%%%%%%%%%%%%%%%%%%%%%%%%%%%%%%%%%%%%%%%%%%%%%%%%%%%%%

% \documentclass{article}
% \usepackage[margin=1in]{geometry}
% \usepackage[utf8]{inputenc}
% % \usepackage{times}
% % \usepackage{helvet}
% % \usepackage{courier}
% \usepackage{xspace}
% \usepackage{algorithm}
% \usepackage{subcaption}
% \usepackage[noend]{algorithmic}
% \usepackage{graphicx} 
% \usepackage{url}
% \usepackage{comment}
% \usepackage{multirow,multicol, makecell, booktabs}
% \usepackage{amsmath}
% \usepackage{amsthm}
% \usepackage{amsfonts}
% \usepackage{color}
% \usepackage{listings}
% \usepackage{todonotes}

% \usepackage{pdflscape}

% \lstset{basicstyle=\scriptsize,
% frame=single
% }
% \frenchspacing
% \setlength{\pdfpagewidth}{8.5in}
% \setlength{\pdfpageheight}{11in}
% \newcommand{\skcomment}[1]{({\color{magenta}SK: #1})}
% \newcommand{\alberto}[1]{\textcolor{blue}{#1}}
% \newcommand{\parisa}[1]{\textcolor{red}{#1}}
% \newcommand{\sarit}[1]{\textcolor{olive}{#1}}
% \newcommand{\francesca}[1]{\textcolor{purple}{#1}}
% \newcommand{\jpmc}{J.P. Morgan\xspace}
% \newcommand{\expres}{\textsc{expres}\xspace}
% \newcommand{\pres}{\textsc{pres}\xspace}

% \title{\textbf{Explaining Preference-driven Schedules: the EXPRES Framework}\\{\Large Supplementary Material}}
% \date{}

% \begin{document}
% \maketitle

\onecolumn
\begin{center}
  {\huge \textbf{Supplementary Material}}  
\end{center}

\vspace{5pt}

\noindent In these Supplementary Material, we report some further details regarding the User Studies 1 and 2 discussed in the main paper. In both studies, we refer to the same two scenarios, depicted in Figures \ref{fig:scenario1} and \ref{fig:scenario2}: in User Study 1 we showed all the preferences and the entire team's schedule; in User Study 2 we showed only one individual set of preferences and one schedule.

\section{Details of User Study 1}
Regarding the User Study 1, that aimed to understand how humans solve \expres tasks, we report the full script of the interviews that were conducted and the list of elicited human-generated explanations.

\subsection{Survey}
\begin{enumerate}
    \item How familiar are you with the Tool and the way it works? [5-point Likert scale anchored with `not familiar at all'-`very familiar']
    \item How often have you interacted with the Tool in the past?
    \item We are going to show you a scenario, which includes the preferences of 8 fictitious employees over a week and the schedule generated by the Tool for that week. \textbf{Why wasn’t Edith assigned to the office on Thursday 18th November, as she requested?}
    Whenever you have come up with your explanation, please type it in the Skype chat of this meeting.
    [See Figure~\ref{fig:scenario1}]
    \item How easy or difficult has it been to provide this explanation? [5-point Likert scale anchored with `very easy'-`very difficult']
    \item We are going to show you another scenario, which includes the preferences of 20 fictitious employees over a week and the schedule generated by the Tool for that week. \textbf{Why wasn’t Ivan assigned to the office on Wednesday 10th November, as he requested?} Whenever you have come up with your explanation, please type it in the Skype chat of this meeting. [See Figure~\ref{fig:scenario2}]
    \item How easy or difficult has it been to provide this explanation? [5-point Likert scale anchored with `very easy'-`very difficult']
\end{enumerate}
 
Between Q2 and Q3, we presented a short summary of the preferences taken into account by the tool when generating a schedule, in order to align the background knowledge of the participants before performing the task.

\subsection{Explanations}
% Here we report all the human-generated explanations that we have collected during the interviews of user study 1.

\paragraph{Scenario 1}
\begin{enumerate}
    \item Other WGs have asked for 18th to be in the office \\ Edith only has 1 - 2 day
    \item Edith wasn't assigned into the office because she hadnt requesred the dates in on the 18th or because since her minimum was 1 and max was 2 and it maxed out the days she can come in
    \item 4 desks on Thursday accounted for 2 people had in office dates \& 2 people had min4 days in and out of office dates on another day meaning that they would be in all other day of the week. 
    \item Working group 1 3 and 3 take precedence for being in the office over Edith's interest in being there on Thursday the 18th.
    \item Too many of the team members have requested to be in the office on a Thursday.
    \item Han was not able to be in the office on that day
    \item What I would tell Edith is: the reason why she was only schedule to come to the office is because she is part working group \# 1 with George and Han. Also she has specify that she would like to work in the office for a max 2 days and a min 2 days where as George and Hans has specify 4-5/3-4 respectively.
    \item She did not introduce that day as a preferred day, nor a days in the office day. Since the remaining constraints for her were satisfied, the generated solution fulfills the constraints
    \item Because of the working group (2) to optimize the number of desks + WG
    \item it is because the other 2 groups were selected to be in on Thursday and Alice group has 3 members that cannot be fit altogether that day
\end{enumerate}

\paragraph{Scenario 2}
\begin{enumerate}
    \item He has asked for preferred day of 8th November, and has min days of 1, so perhaps the AI will only allocate 1 to Ivan to accommodate others going to the office more?
    \item Ivan was not requested to come in on the 10th because he did not ask or coordinate with the scheduler person to assign him in the ""dates in office"" section to come in on that specific date or because 
    of the working groups and the assigned desks given to the team
    \item 3 people had dates in, another 2 had 5 days min, the rest are perhaps a combination of all the other factors, meaning they had to be in instead of Ivan
    \item Ivan was not assigned to go into the office on Wednesday the 10th because preferred days is the last condition that is satisfied
    \item Due to being a large group and many of the employees requesting the same day, not everyone could get their desired day in the office.
    \item He has 1 min days per week
    \item The reason why I think he wasn't schedule those two days is because there are currently 12 desk available and the tool have already schedule many teams on that specific day and it become very difficult to accommodate Tuesday \& Wednesday
    \item He marked that day as a preferred day. Since this type of constraint is the least priority one, the solution could not accommodate the constraints of everyone plus Ivan´s
    \item To optimize the working groups in the office + the number of desks
    \item He was limited by the seat availability that day , members of two other groups has selected the 1th specifically to be in so they took preference 
\end{enumerate}

\begin{landscape}
\begin{figure}
    \centering
    \includegraphics[width=\linewidth]{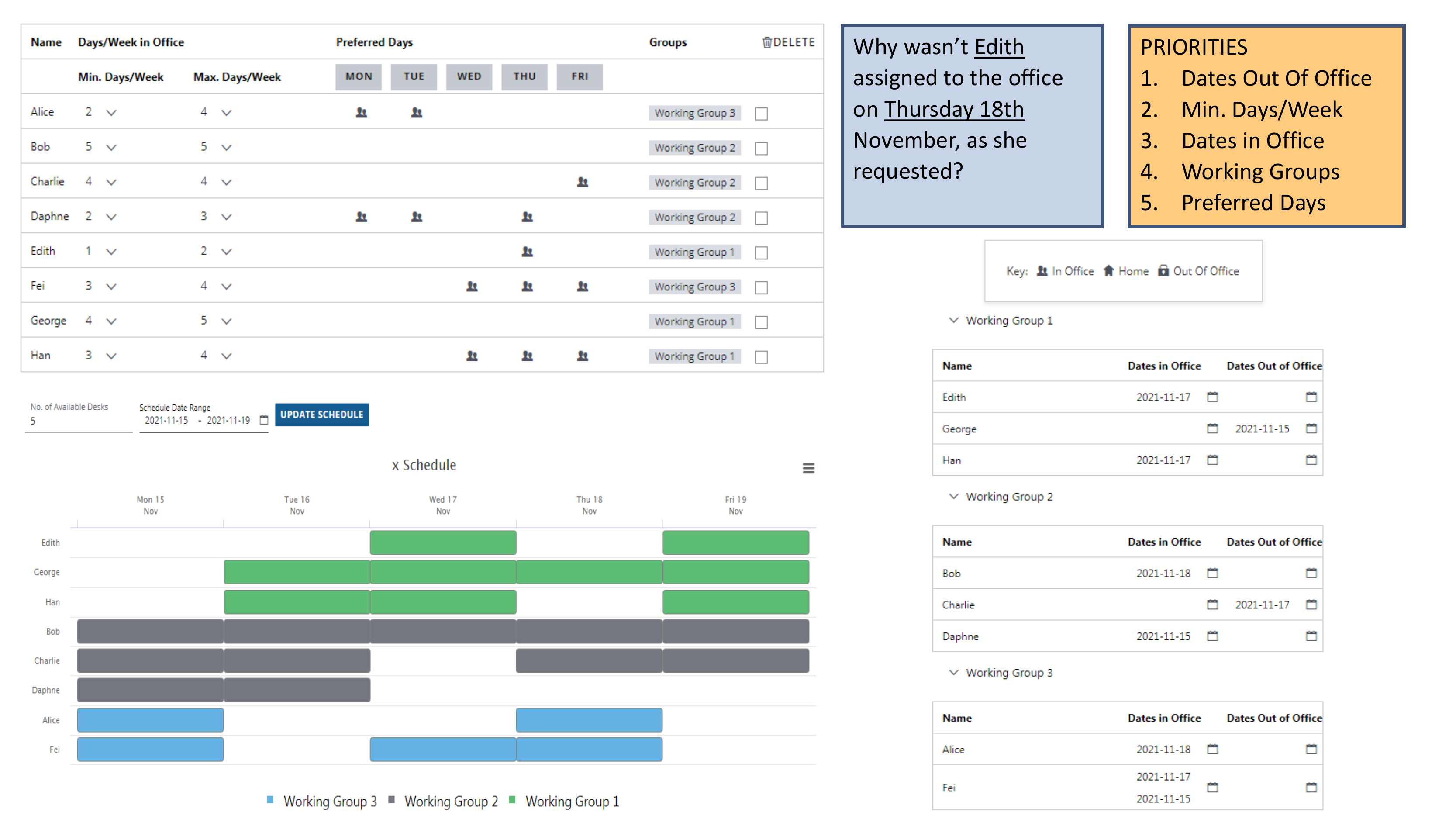}
    \caption{Scenario 1}
    \label{fig:scenario1}
\end{figure}

\begin{figure}
    \centering
    \includegraphics[width=\linewidth]{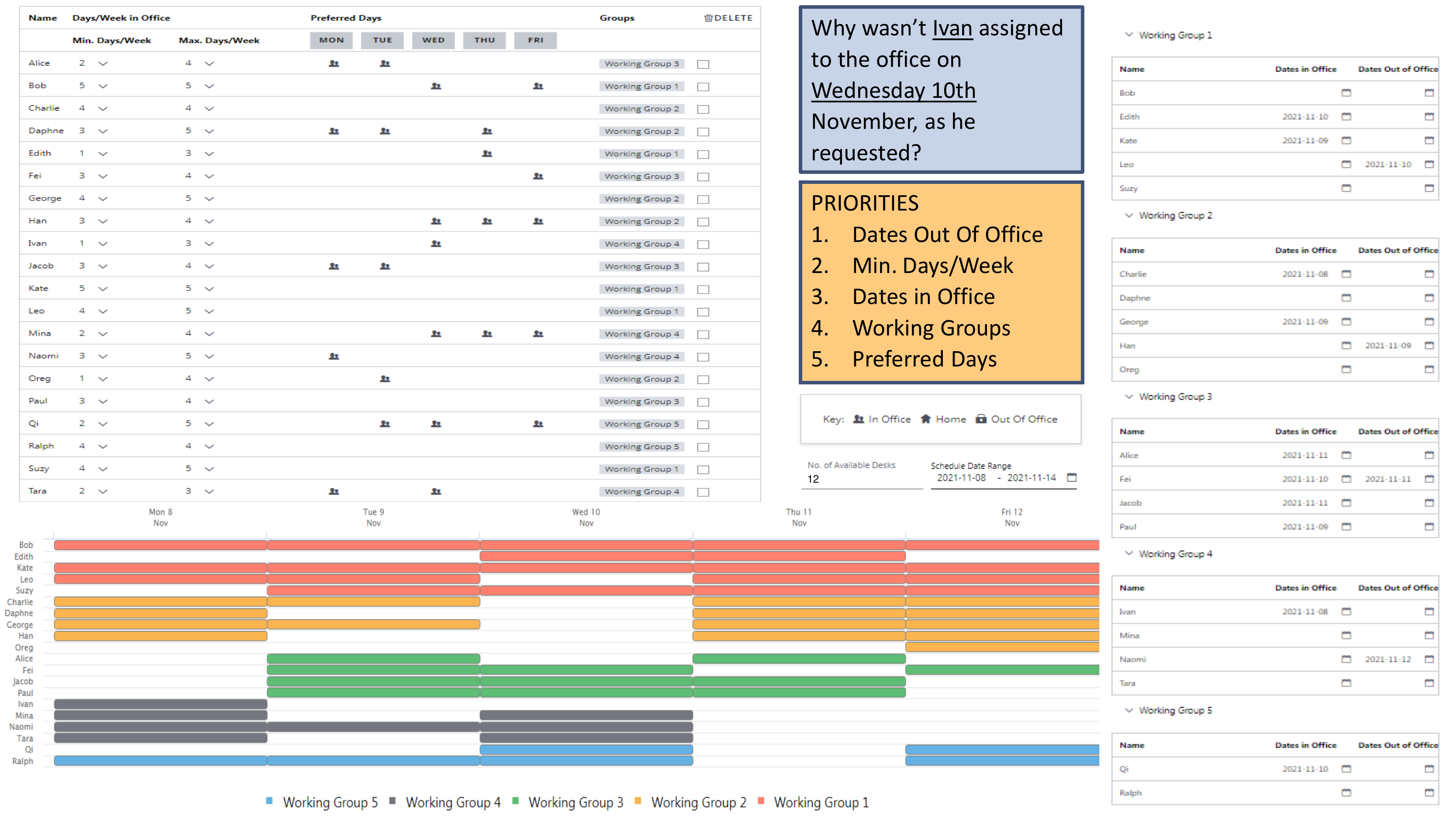}
    \caption{Scenario 2}
    \label{fig:scenario2}
\end{figure}
\end{landscape}

\newpage
\section{Details of User Study 2}
Regarding the User Study 2, that aimed to understand which type of explanation (\expres-generated or human-generated) was more preferred, we report the full questionnaire that was presented to the participants.

Note that we selected the explanations to include in this study as follows. 
Regarding the human-generated ones from User Study 1, three people independently shortlisted the three explanations they considered of the highest quality; then, through majority voting, with a fourth person involved to break the ties, the ones we report below were selected. 
Regarding the EXPRES-generated explanations, we aimed to maximise the diversity of reason-types appearing in the explanations (cf. Table 4 in the main paper).

\subsection{Scenario 1}
Imagine you are Edith and you work in a team of 8 people. In particular, you collaborate closely with other two employees.

In order to better organize the assignment of desks at the office, you shared the following preferences with your team manager:
\begin{itemize}
    \item Minimum number of day/week: 1
    \item Maximum number of days/week: 2
    \item Meetings (dates in office): 17 Nov 2021
    \item Working group: George and Han
    \item Preferred day: Thursday
\end{itemize}

A tool automatically generates an optimal schedule by taking into account individual preferences according to the following priorities:
\begin{enumerate}
    \item Min. Days/Week
    \item Meetings (dates in office)
    \item Working Groups
    \item Preferred Days
\end{enumerate}
Meaning that first it tries to satisfy all the minimum number of days/week, then the meetings (dates in office), the working groups and at last the individually preferred days.
 
You (Edith) receive the following schedule:

\bigskip
\begin{tabular}{c|c|c|c|c|c}
    & Mon 15 Nov &	Tue 16 Nov &	Wed 17 Nov &	Thu 18 Nov &	Fri 19 Nov \\
    \hline
    Edith & & & in office & & in office
\end{tabular}

\paragraph{Question 1}
Why wasn't your (Edith's) preference for Thursday respected?
Please select the 3 explanations that satisfy you most. 
\textit{Note that these were presented in a random order.}

\begin{itemize}
    \item[E1] The preference could not be satisfied because the 5 available desks were assigned to other people with more important preferences: 3 employees due to minimum number of days per week; 1 employee due to meetings; 1 employee due to 1 working group.
    \item[E2] The preference could not be satisfied because the 5 available desks were assigned to other people with more important preferences: 1 employee due to minimum number of days per week; 4 employees due to 2 working groups. 
    \item[E3] The preference could not be satisfied because the 5 available desks were assigned to other people with more important preferences: 1 employee due to minimum number of days per week; 2 employees due to meetings; 1 employee due to working group; 1 employee due to preferred day. 
    \item[H4] Too many of the team members have requested to be in the office on a Thursday.
    \item[H5] Edith wasn't assigned into the office because she hadnt requesred the dates in on the 18th or because since her minimum was 1 and max was 2 and it maxed out the days she can come in
    \item[H6] 4 desks on Thursday accounted for 2 people had in office dates \& 2 people had min4 days in and out of office dates on another day meaning that they would be in all other day of the week. 
\end{itemize}

\paragraph{Question 2}
Why wasn't your (Edith's) preference for Thursday respected?
Please rank the explanations that you previously selected according to your satisfaction with them (1=most satisfying, 3=least satisfying). \textit{Note that in these question only the three previously selected explanations were shown.}

\subsection{Scenario 2}
Imagine you are Ivan and you work in a team of 20 people. In particular, you collaborate closely with other three employees.

In order to better organize the assignment of desks at the office, you shared the following preferences with your team manager:
\begin{itemize}
    \item Minimum number of day/week: 1
    \item Maximum number of days/week: 2
    \item Meetings (dates in office): 8 Nov 2021
    \item Working group: Mina, Naomi and Tara
    \item Preferred day: Wednesday
\end{itemize}

A tool automatically generates an optimal schedule by taking into account individual preferences according to the following priorities:
\begin{enumerate}
    \item Min. Days/Week
    \item Meetings (dates in office)
    \item Working Groups
    \item Preferred Days
\end{enumerate}
Meaning that first it tries to satisfy all the minimum number of days/week, then the meetings (dates in office), the working groups and at last the individually preferred days.

\bigskip
\begin{tabular}{c|c|c|c|c|c}
    & Mon 15 Nov &	Tue 16 Nov &	Wed 17 Nov &	Thu 18 Nov &	Fri 19 Nov \\
    \hline
    Ivan & in office & &  & & 
\end{tabular}

\paragraph{Question 3}
Why wasn't your (Ivan's) preference for Wednesday respected?
Please select the 3 explanations that satisfy you most.
\textit{Note that these were presented in a random order.}

\begin{itemize}
    \item[E1] The preference could not be satisfied because the 12 available desks were assigned to other people with more important preferences: 3 employees due to minimum number of days per week; 2 employees due to meetings; 7 employees due to 3 working groups. 
    \item[E2] The preference could not be satisfied because the 12 available desks were assigned to other people with more important preferences: 12 employees due to 4 working groups.
    \item[E3] The preference could not be satisfied because the 12 available desks were assigned to other people with more important preferences: 3 employees due to minimum number of days per week; 8 employees due to 4 working groups; 1 employee due to preferred day.
    \item[H4] Due to being a large group and many of the employees requesting the same day, not everyone could get their desired day in the office.
    \item[H5] 3 people had dates in, another 2 had 5 days min, the rest are perhaps a combination of all the other factors, meaning they had to be in instead of Ivan
    \item[H6] He was limited by the seat availability that day , members of two other groups has selected the 1th specifically to be in so they took preference 
\end{itemize}

\paragraph{Question 4}
Why wasn't your (Ivan's) preference for Wednesday respected?
Please rank the explanations that you previously selected according to your satisfaction with them (1=most satisfying, 3=least satisfying). \textit{Note that in these question only the three previously selected explanations were shown.}

% \end{document}

\end{document}